\documentclass{article}
\usepackage{ijcai22}

\usepackage{times}

\usepackage{soul}
\usepackage{url}
\usepackage[hidelinks]{hyperref}
\usepackage[utf8]{inputenc}
\usepackage[small]{caption}
\usepackage{graphicx}
\usepackage{amsmath}
\usepackage{booktabs}
\usepackage{algorithm}
\usepackage{algorithmic}
\usepackage{multirow}
\usepackage{subfigure}
\title{A Closed-Loop Perception, Decision-Making and Reasoning Mechanism for Human-Like Navigation}

\author{
Wenqi Zhang$^{1}$\footnotemark[1]\and
Kai Zhao$^{2}$\footnotemark[1]\and
Peng Li$^{3,6}$\and
Xiao Zhu$^4$\and  \\
Yongliang Shen$^1$\and
Yanna Ma$^5$\and
Yingfeng Chen$^2$\And
Weiming Lu$^{1}$\footnotemark[2]\\
\affiliations
$^1$College of Computer Science and Technology, Zhejiang University\\
$^2$Netease Fuxi Robot Department\\
$^3$Institute of Software, Chinese Academy of Sciences\\
$^4$College of Mechanical Engineering, Zhejiang University of Technology\\
$^5$University of Shanghai for Science and Technology\\
$^6$University of Chinese Academy of Sciences Nanjing\\
\emails
{\{zhangwenqi, luwm\}@zju.edu.cn},\{zhaokai02,chenyingfeng1\}@corp.netease.com, lipeng@iscas.ac.cn}

\begin{document}
\maketitle

\renewcommand{\thefootnote}{\fnsymbol{footnote}} 
\footnotetext[1]{Contributed equally to this work. $^\dag$Corresponding author.}  
\renewcommand{\thefootnote}{\arabic{footnote}}

\begin{abstract}
Reliable navigation systems have a wide range of applications in robotics and autonomous driving. Current approaches employ an open-loop process that converts sensor inputs directly into actions. However, these open-loop schemes are challenging to handle complex and dynamic real-world scenarios due to their poor generalization. Imitating human navigation, we add a reasoning process to convert actions back to internal latent states, forming a two-stage closed loop of perception, decision-making, and reasoning. Firstly, VAE-Enhanced Demonstration Learning endows the model with the understanding of basic navigation rules. Then, two dual processes in RL-Enhanced Interaction Learning generate reward feedback for each other and collectively enhance obstacle avoidance capability. The reasoning model can substantially promote generalization and robustness, and facilitate the deployment of the algorithm to real-world robots without elaborate transfers. Experiments show our method is more adaptable to novel scenarios compared with state-of-the-art approaches. 

\end{abstract}
\section{Introduction}

Safe, reliable, and flexible navigation and obstacle avoidance strategies are essential for large-scale robotic and autonomous driving applications. In a complex scenario of human-robot coexistence, the agent needs to avoid highly dynamic obstacles and drive quickly to the target, which is fairly challenging. Most conventional algorithms\cite{fox1997dynamic} implement navigation through real-time path planning, which requires a high computational overhead.

Recent advances in deep learning have greatly improved perception capabilities and even surpassed human levels in many areas\cite{he2016deep,vaswani2017attention}. But supervised learning paradigm, which relies on large amounts of training data, does not easily tackle complex navigation problems in highly dynamic scenarios. Navigation is inherently a complex sequential decision-making task, including global path planning and local obstacle avoidance. It is impractical to collect sufficient data for training. Obviously, Complex decision-making tasks in navigation and obstacle avoidance cannot be well solved by perception model alone.

\begin{figure}[t]

	\centering
	\includegraphics[width = 1\linewidth]{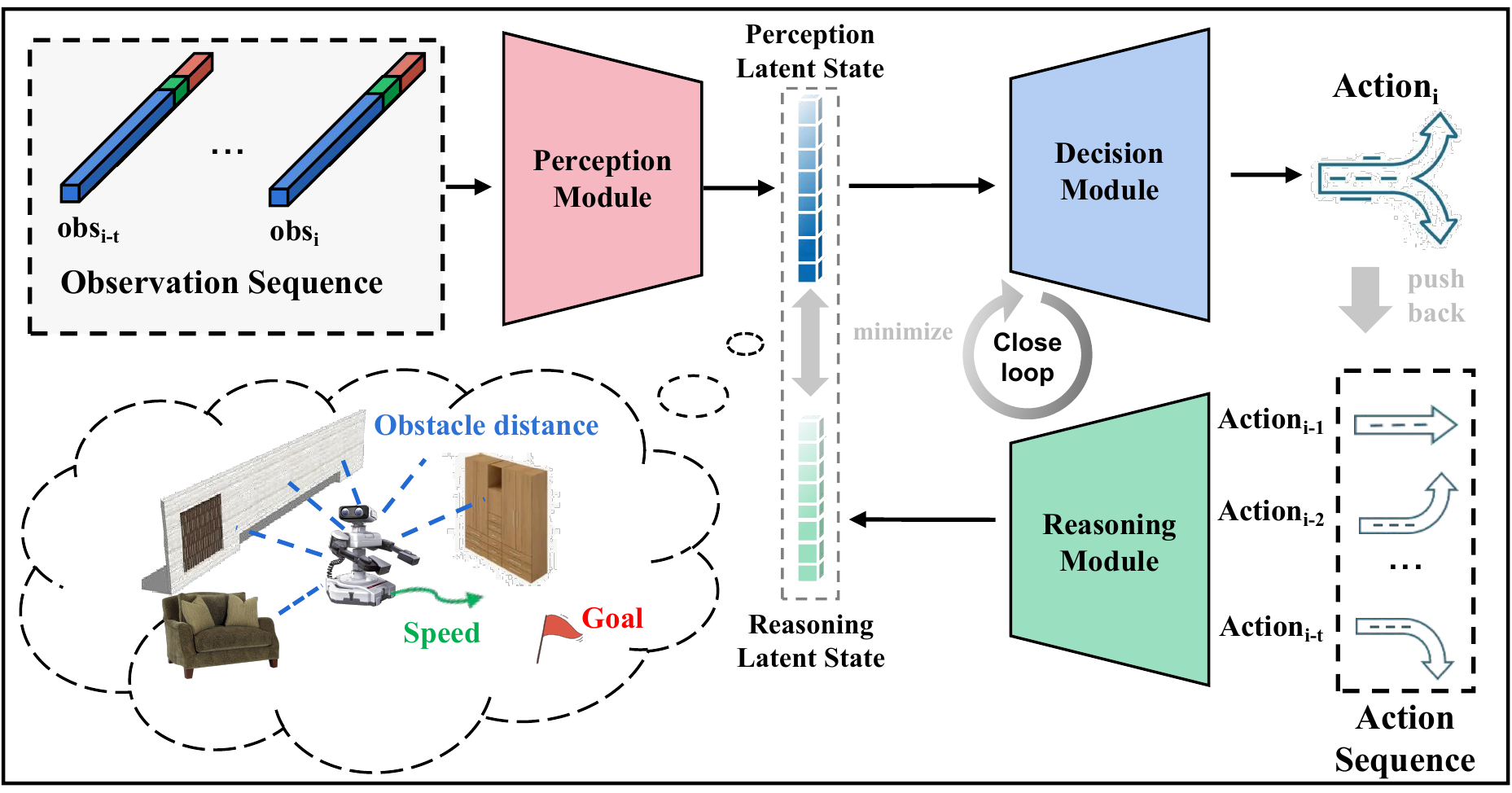}

\caption{We propose a closed-loop mechanism including perception, decision-making, and reasoning model. The perception model extracts a latent representation of both spatial and temporal aspects of observation. The decision-making model calculates the most appropriate action according to the latent state. The reasoning model imagines this latent representation of the surroundings through the action sequences in memory.}
	\label{fig:demo}
\end{figure}

Deep reinforcement learning(DRL)\cite{mnih2013playing,sutton2018reinforcement} has been widely studied and applied in many fields, including robot control\cite{yang2020multi,fu2021deep}, autonomous driving\cite{long2018towards}, etc. The DRL-based approach shows great potential to build a reliable decision-making system. Currently, many researchers have applied DRL to navigation and obstacle avoidance\cite{faust2018prm,vuca}, achieving promising performance. Most of these works attempt to learn a precise policy from observations to actions during environment interactions. However, such one-way learning process 
is open-loop and lacks a deep understanding of the action. These open-loop methods perform well in trained scenarios, but may not be robust in general scenarios without understanding the intrinsic relations between internal states and output action. Besides, DRL-based methods are fragile and unstable when deployed to real scenarios, since the real world is dynamic, complex, and always changing. 







However, for the person with visual impairment, we observe a reasoning process in their mind. They perceive the outside obstacles through their crutch and imagine the surroundings based on their motion when exploring an unfamiliar scene. The imaginary scenes constructed in their minds match the real scenes in reality, forming a closed loop. Inspired by this, we design a two-stage closed-loop mechanism with perception, decision-making, and reasoning components. As shown in Figure \ref{fig:demo}, the perception model is to compress observation into latent state distribution and the decision-making model converts the latent state into current action, just as a blind person uses a crutch to explore a new scene. Simultaneously the reasoning model deduces the most likely latent states based on the action sequences, like the person imagining the surrounding scene in their mind. This reasoning process constructs a closed-loop learning process by parsing the intrinsic relation between actions and the latent states, generating more reasonable actions than the open-loop system. Unlike conventional feedback control algorithms, our algorithm focuses on the closed loop of the learning process, eliminating the need to accurately model the kinematics and dynamics of the
robot and the environment.




%



We assess our algorithm on two benchmarks that we design for evaluating navigation ability in few-shot and zero-shot scenes. Experiments demonstrate that our approach achieves signiﬁcant improvement over various baselines, and is more reliable in novel scenarios. In addition, we deploy the algorithm to a real robot in a crowded building. Despite the significant gap between the simulator and the real world, the agent achieves autonomous obstacle avoidance without any human assistance. 
The main contributions of this work can be summarized as follows:


\begin{itemize}
\item We propose a perception, reasoning, and decision-making mechanism, which can learn more general rules and robust strategies through the reasoning model and latent state distribution.


\item We introduce VAE-Enhanced Demonstration Learning to acquire basic navigation rules from data and design two dual-learning processes in RL-Enhanced Interaction Learning to promote collision avoidance ability by generating reward feedback for each other.
\item Extensive experiments show our method achieves the most reliable results, surpassing the previous approaches in terms of safety and task accomplishment.
\end{itemize}

\section{Related Work}
\subsection{Navigation and Planning}
In recent years, autonomous navigation and flexible obstacle avoidance have attracted extensive attention. Conventional navigation approaches\cite{khatib1986real} are usually implemented by global and local path planning\cite{fox1997dynamic}. These approaches have trouble addressing navigation challenge in an unknown environment as they must be provided with the map in advance. In addition, many dynamic path planning and re-planning algorithms\cite{stentz1995focussed,koenig2002d,9115288}  can also achieve path planning in unknown dynamic environments, but the high computational overhead of these algorithms constrains real-time performance in large-scale dynamic environments. Besides, most of them require fine-tuning of parameter in real applications.

With the breakthrough of deep learning, many researchers have  deployed supervised learning and reinforcement learning to achieve automatic obstacle avoidance\cite{chen2017socially}. But collecting abundant samples for training is another annoying issue, as manual annotation requires tremendous tedious labor. To overcome this limitation, some researchers attempt to adopt imitation learning\cite{ho2016generative,pfeiffer2018reinforced}. Imitation algorithm designs a teacher-student learning paradigm to mimic the teacher's action output and the intention behind it. Furthermore, some researches focus on modeling the environment. \cite{ha2018recurrent} adopted a variational autoencoder(VAE) to learn a compressed spatial and temporal representation of the environment. Although the above-mentioned studies have strikingly improved the performance of the navigation, most of them are based on learning an explicit mapping from perfect observations to deterministic actions. Such an assumption is impractical in a real scenario. These methods may have weak robustness and generalization, especially when confronted with novel scenarios.

\subsection{Deep Reinforcement Learning}
Deep Reinforcement Learning(DRL) is an optimization algorithm based on Markov Decision Process(MDP). Its optimization goal is to maximize the expected accumulated discount rewards\cite{sutton2018reinforcement,mnih2013playing}. Many researchers have applied DRL to traditional decision-making tasks and achieved remarkable results. \cite{tsounis2020deepgait} applied DRL to robot control and realized autonomous walking of a quadruped robot. \cite{long2018towards,vuca} introduced DRL to solve path planning and obstacle avoidance in a dynamic environment. Some low-resource 
tasks\cite{NIPS2016_5b69b9cb,artetxe2017unsupervised,cao2020unsupervised} can also be solved using dual DRL. These dual-DRL frameworks utilize the dual property of two tasks to enhance the model performance. However, DRL also has some inherent drawbacks that hinder its performance. One of the troubling challenges is commonly referred to as the “curse of dimensionality”, which hinders the collision avoidance performance in large-scale real scenarios. The other is the poor generalizability and robustness when encountering  complex and dynamic scene.



\section{Approach}
In this section, we first introduce the perception, reasoning, and decision-making components, and then describe in detail the design of VAE-Enhanced Demonstration Learning and RL-Enhanced Interaction Learning.
\subsection{Model Definition}
In previous approaches, the model directly maps external observations into actions \cite{long2018towards,vuca}. In fact, these observations also contain lots of irrelevant information and measurement noise. We use $o_{t}$ to denote the current observation, $a_{t}$ for current action, and $s_{t}$ for current latent state based on observation. Assuming that $s_{t}$ obeys a normal distribution $N_{t}(s)$ and it represents the valid features extracted from $o_{t}$. The perception, decision-making and reasoning models achieves the transition between $o_{t}$, $a_{t}$, $s_{t}$. The perception and reasoning are two sequence encoders based on observations and actions, while decision-making is a simple decoder depending on the current latent state. This process is similar to human navigation that the perception and reasoning are based on long-term memory, while decision making is an instant subconscious response. 

\paragraph{Perception component.} The perception model is a sequence encoder $P$ that maps observations sequences $o_{t-n:t}$ into the latent state distribution $N_{t}^{P}$, where $o_{t-n:t}=o_{t-n} \cdots o_{t}$ denotes the sequences of observation, and latent state is sampled from latent state distribution, i.e. $s_{t}^{P} \sim N_{t}^{P}$. The perception model extracts useful features from the observed sequences, and compresses them into the latent state, filtering out the irrelevant information.

\paragraph{Decision-Making component.} The decision model is a direct mapping from latent states to actions, i.e. $a_{t}=D\left(s_{t}\right)$.

\paragraph{Reasoning component.} The reasoning model $R$ is to deduce the latent state distribution $N_{t}^{R}$ based on the action sequence $a_{t-n:t}$, where $a_{t-n:t}=a_{t-n} \cdots a_{t}$ represent the sequences of actions. Another latent state is sampled from this distribution, i.e. $s_{t}^{R} \sim N_{t}^{R}$. This reasoning process is similar to a person imagining the surrounding scenes based on movements.

\subsection{VAE-Enhanced Demonstration Learning}

\begin{figure}[t]

	\centering
	\includegraphics[scale=0.8,width = 1\linewidth]{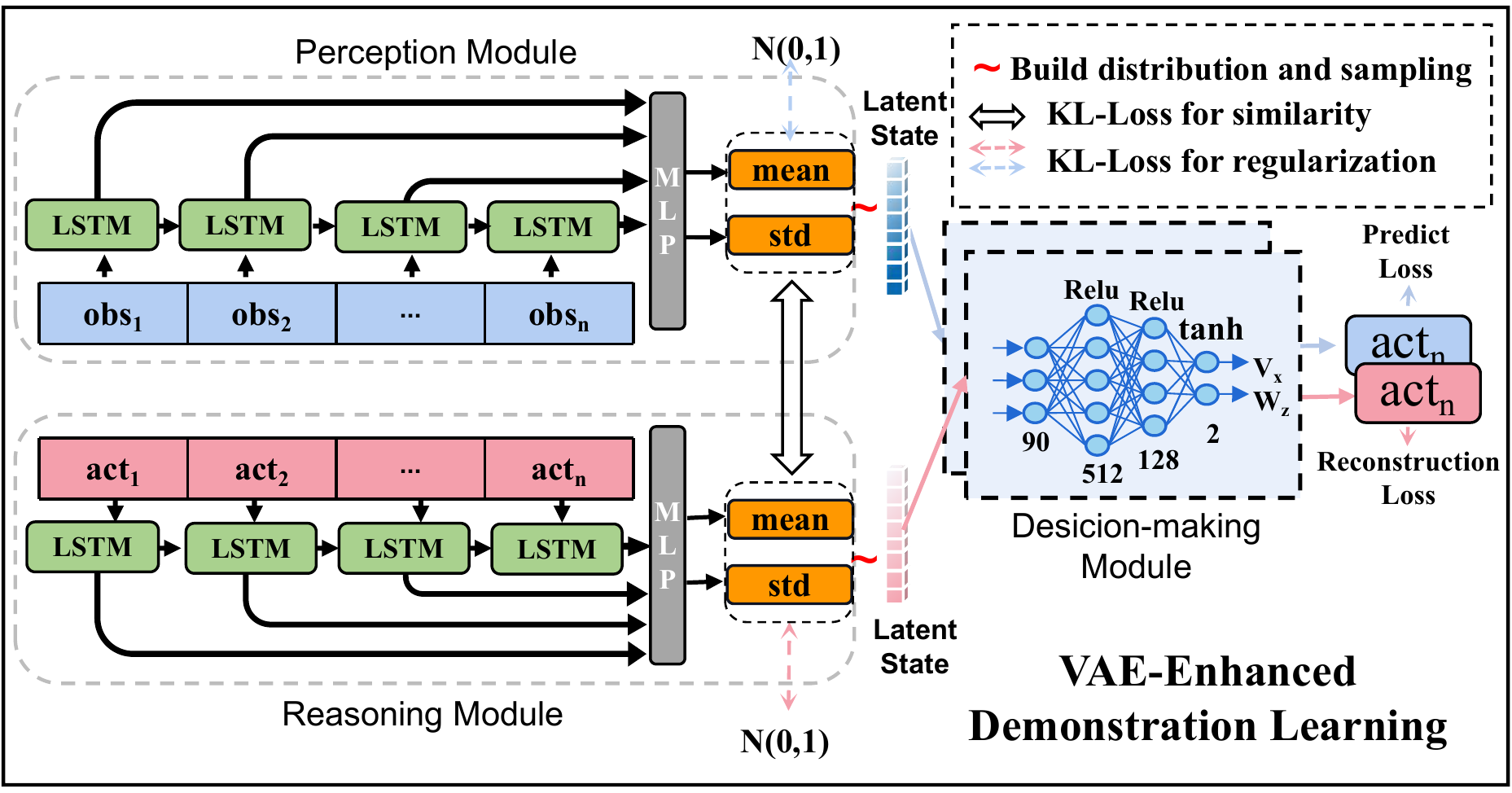}

\caption{Overview of the VAE-Enhanced Demonstration Learning.}
	\label{fig:first_stage1}
\end{figure}

Using DRL to directly train navigation policy from scratch is inefficient since DRL usually encounters a cold start problem. We design a VAE-Enhanced phase to learn from data. This process is similar to imitation learning, where three models learn the general rules from the data. The overview of this process is visualized in Figure \ref{fig:first_stage1}. It contains two channels: 
\begin{itemize}
\item \emph{Prediction-Denoise-channel}: decision-making (D) and perception model(P)

\item \emph{Reconstruction-VAE-channel}: decision-making(D) and reasoning model(R)

\end{itemize}

\paragraph{Prediction-Denoise-channel.} Pair data$ {\left\{o_{t},a_{t} \right\}}_{t=1:N}$ sampled from collected dataset (See section 4.1 for more detailed descriptions about dataset). It means that when $o_{t}$ is observed, the agent should take $a_{t}$. $p(s_{t}|o_{t-n:t})$ means perception model and $p(a_{t}|s_{t})$ represents decision-making model. Latent state cannot be observed. The likelihood as follows:
\begin{small}
\begin{equation}
p(a | o) =\int p(s, a | o)ds=\int p(a | s) p(s | o) ds=\int D(\cdot)P(\cdot)ds
\end{equation}%
\end{small}

The algorithm maximizes the log-likelihood of the data $\{o_{t},a_{t}\}$. So, we adopt a \emph{Prediction-Denoise-channel} to achieve this objective. The perception model converts observation sequences $o_{t-n:t}$ into latent distribution $N^{P}_{t}(s)$, and then the decision-making model converts latent state with sampling noise $s_{t}^{P}$ to action $a_{t}$. The perception and decision-making models are more robust for predicting actions due to the presence of sampling noise in latent state. The \emph{prediction-channel} can be formalized as
\begin{align}
&\mu_{t}^{P},\sigma_{t}^{P}=\operatorname{MLP}^{P}(\operatorname{LSTM}^{P}(o_{t-n:t}))\\ 
&s_{t}^{P} \sim N(\mu^{P}_{t}, \sigma^{P}_{t})\\
&a_{t}^{pred}=\operatorname{MLP}^{D}(s_{t}^{P})
\end{align}      

\paragraph{Reconstruction-VAE-channel.} Inspired by \cite{kingma2013auto}, we adopt a \emph{Reconstruction-VAE-channel} to extract a continuous mapping for reasoning model. Assume that true distribution from action to latent state is $p^{*}(s|a)$, and the reasoning model $R(s_{t}|a_{t-n:t})$ achieve an estimate of this distribution. To minimize the KL divergence of the two distributions:
\begin{align}\nonumber
&KL\left(R\left(s|A\right)\| p^{*}(s | a)\right)=\int R(s|A)  \log \frac{R(s|A)}{p^{*}(s | a)} d s \\\nonumber
&=\int R(s|A) \log \frac{R(s|A)}{p^{*}(s)} d s-\int R(s|A)\log p(a | s) d s \\\nonumber
&=K L(R(s | A) \| p^{*}(s))-E_{s \sim R(s|A) }[\log p(a | s)]\\ 
&=KL(R(\cdot)\|p^{*}(s))-E_{s \sim R(\cdot) }[\log D(\cdot)]
\end{align}

\begin{figure}[t]
	\centering
	\includegraphics[width = 1\linewidth]{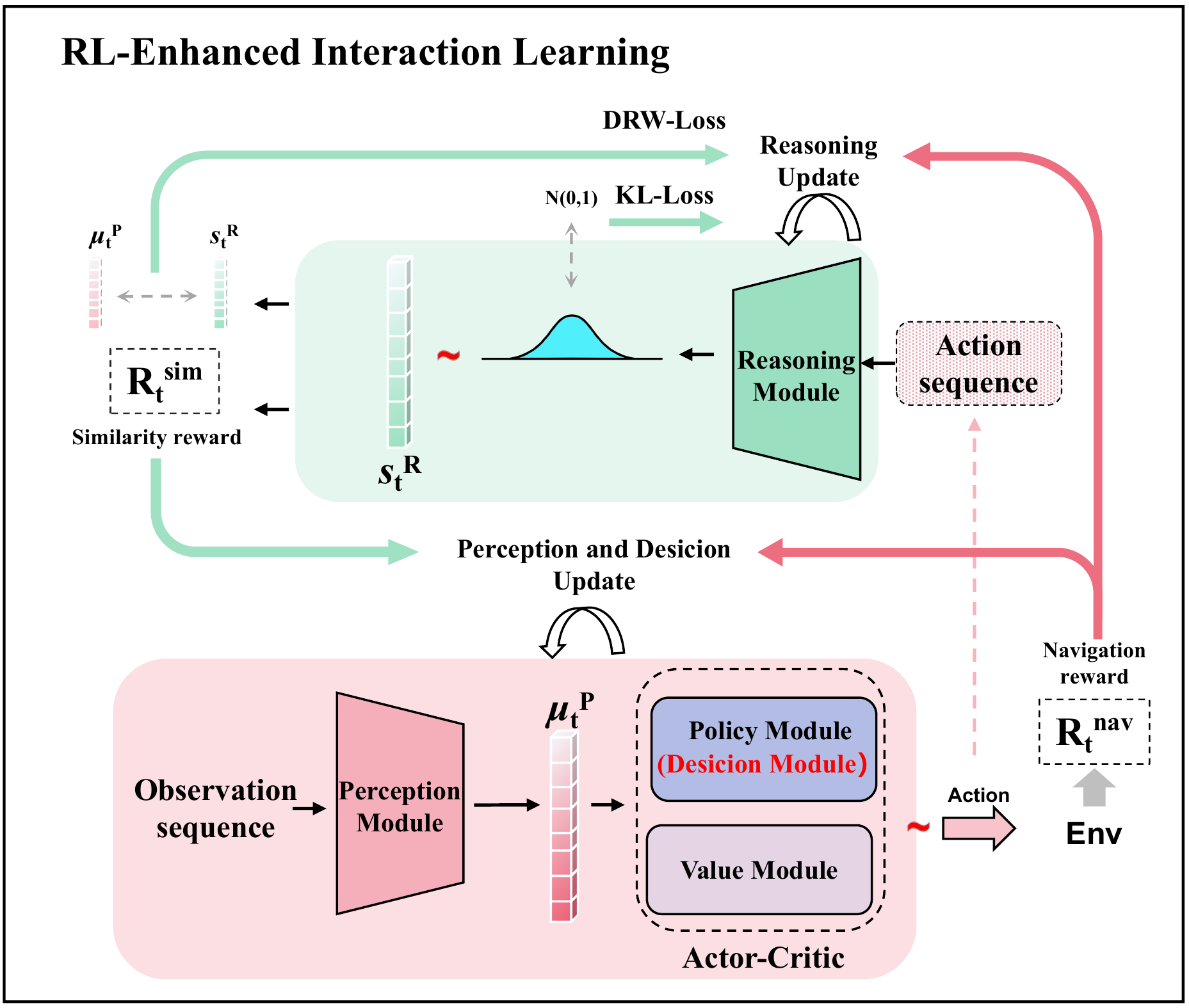}
\caption{The reasoning model updates itself at a fixed interval by selecting high navigation-reward actions. Meanwhile, the reasoning model generates a similarity-reward to evaluate each action generated by the decision-making model during the interaction.}
	\label{fig:second_stage}
\end{figure} 
Where A means action sequence. The reasoning and decision-making models constitute a standard VAE process, which achieves the reconstruction of the action. Given an action sequence, reasoning model encodes $a_{t-n:t}$ into latent state distribution $N_{t}^{R}$, i.e. $\mu_{t}^{R},\sigma_{t}^{R}=\operatorname{MLP}^{R}(\operatorname{LSTM}^{R}(a_{t-n:t}))$. Decision-making model samples another latent state $s_{t}^{R}$, and then reconstructs action $a_t$, denoted as $a_{t}^{reconst}=\operatorname{MLP}^{D}(s_{t}^{R})=\operatorname{MLP}^{D}(s \sim N^{R}_{t})$. 

Two channels share one decision-making model, achieving action prediction and reconstruction, respectively. The training objective can be formulated as
\begin{align}
L_{1,t}&= (a^{reconst}_{t}-a_{t})^{2} + (a^{pred}_{t}-a_{t})^{2}\\
L_{2,t}&=\operatorname{JS}(N(\mu^{P}_{t},\sigma^{P}_{t}) \| N({\mu}^{R}_{t}, \sigma^{R}_{t}))\\
L_{3,t}&=\sum_{\Delta \in \{P,R\}}\frac{1}{2}(-1-\log \sigma_{t}^{\Delta}+(\mu_{t}^{\Delta})^{2}+\sigma_{t}^{\Delta})
%
%
\end{align}

The total loss is  
$L=\sum_{t}(L_{1,t}+L_{2,t}+L_{3,t})$, where $t=1,2 \cdots K$, $K$ is the total number of samples and JS stands for Jensen–Shannon divergence. $L_{1}$ represents the prediction loss and reconstruction loss for two channels. $L_{2}$ constrains that the latent state distribution calculated by the perception and reasoning models should be similar. $L_{3}$ is the regularization constraints two latent distributions to the standard normal distribution. The introduction of the VAE and two channels enhance the generalization and robustness of the three models, avoiding overfitting sparse data.

\subsection{RL-Enhanced Interaction Learning}
After VAE-Enhanced Demonstration Learning, the model is able to learn general collision avoidance rules. However, the limited training data still confines the performance of the algorithm in highly dynamic scenarios. We introduce the RL-Enhanced Interaction Learning to enhance dynamic obstacle avoidance capability through two dual learning processes. 

The perception and decision-making models explore the reasonable actions by interacting with the environment, while the reasoning model deduces the latent state distribution based on the action sequence. These two dual processes generate reward feedback for each other and collectively enhance their performance. The whole process is similar to a blind person carefully exploring the outdoor scene and always imagining the surrounding according to the movements.

As shown in Figure \ref{fig:second_stage}, the trained parameters from the first phase are used to initialize three models. A value network with random initialization and decision-making model form an Actor-Critic framework for proximal policy optimization(PPO)\cite{schulman2017proximal} update. 

First, the current $o_{t}$ and historical observation sequences $o_{t-n:t-1}$ are concatenated together and fed into the perception model to calculate the mean $\mu_{t}^{P}$ of the latent state distribution. Since policy module in Actor-Critic involves a sampling process, we directly treat $\mu_{t}^{P}$ as the latent state $s_{t}$ without sampling. Then decision-making model(policy module) predicts the action $a_{t}^{pred}$ based on $s_{t}$. After that, the environment executes the action $a_{t}^{pred}$ and returns \emph{navigation-reward $R^{nav}_{t}$} . Simultaneously, the reasoning model infers the most likely distribution of latent state $N_{t}^{R}$ based on the current action $a_{t}^{pred}$ and the historical real actions $a_{t-n:t-1}^{real}$. The probability of $\mu_{t}^{P}$ in $N_{t}^{R}$ is treated as a \emph{similarity-reward $R^{sim}_{t}$} for evaluating the quality of the $a_{t}$. This process can be formulated as
\begin{small}
\begin{align}
&\mu_{t}^{P} =\operatorname{MLP}^{P}(\operatorname{LSTM}^{P}([o_{t};o_{t-n:t-1}]))\\ &a_{t}^{pred}=\operatorname{MLP}^{D}(\mu_{t}^{P})\\
&o_{t+1},R_{t}^{nav} = Env(a_{t}^{pred})\\
&\mu_{t}^{R}, \sigma_{t}^{R}=\operatorname{MLP}^{R}(\operatorname{LSTM}^{R}([a_{t}^{pred};a^{real}_{t-n:t-1}]))\\ 
&R_{t}^{sim}=\operatorname{Prob} (\mu^{P}_{t}\mid N(\mu^{R}_{t}, \sigma^{R}_{t})) 
\end{align}
\end{small}

\paragraph{Optimization for perception and decision-making.} 
The PPO is employed to optimize the perception and decision-making model via maximizing the expected cumulative reward($R^{nav}_{t}+R^{sim}_{t}$). The loss functions of perception and decision-making model are calculated as: $L^{P, D}= \alpha*L^{policy}+\beta*L^{value}-\eta*H$, where $H, L^{policy}, L^{value}$ represent the policy entropy, policy loss, and value loss in PPO update, respectively, and $\alpha,\beta,\eta$ are hyper-parameters.

\begin{table*}[h]
\centering
\resizebox{\linewidth}{!}{

\renewcommand{\arraystretch}{1.2}
\setlength\tabcolsep{2pt}
\begin{tabular}{ccrrrrrrrrrr}
\toprule
\begin{tabular}[c]{@{}c@{}}Zero-shot \\ scene benchmark\end{tabular}                                                                                             &  \multicolumn{1}{c}{\begin{tabular}[c]{@{}c@{}}Metrics\\mean/std \end{tabular}} & \multicolumn{1}{c}{\begin{tabular}[c]{@{}c@{}}S$_{1}$+Density\\ change  \end{tabular}} & \multicolumn{1}{c}{\begin{tabular}[c]{@{}c@{}}S$_{2}$+Shape\\change  \end{tabular}} & \multicolumn{1}{c}{\begin{tabular}[c]{@{}c@{}}S$_{3}$+Density\\change  \end{tabular}} & \multicolumn{1}{c}{\begin{tabular}[c]{@{}c@{}}S$_{4}$+Speed\\ change\end{tabular}} & \multicolumn{1}{c}{\begin{tabular}[c]{@{}c@{}}S$_{5}$+Volume\\change \end{tabular}} & \multicolumn{1}{c}{\begin{tabular}[c]{@{}c@{}}S$_{6}$+
Obstacle\\change \end{tabular}} & \multicolumn{1}{c}{\begin{tabular}[c]{@{}c@{}}S$_{7}$+Edge\\change\end{tabular}} & \multicolumn{1}{c}{\begin{tabular}[c]{@{}c@{}}S$_{8}$+View\\change\end{tabular}} & \multicolumn{1}{c}{AVG} \\ \midrule
\multicolumn{11}{c}{One-stage method}    \\ \midrule  
Move-base                                                                      & SR      & 60\%                                                                                 & 70\%                                                                                    & 65\%                                                                                     & 10\%                                                                                              & 75\%                                                                                                      & 15\%                             & 65\%                                                                                             & 30\%                                                                                                               & 48.75\%                         \\ 
                                                                                              
\multirow{2}{*}{DM-RCA}                                                                          & SR      & 47.50\%(3.7)                                                                          & 69.75\%(4.5)                                                                             & 59.25\%(3.3)                                                                              & 48.25\%(3)                                                                                              & 78\%(4.6)                                                                                                 & 62.75\%(6.9)                             & 87.75\%(4.7)                                                                                      & 73.5\%(5)                                                                                                          & 65.84\%                        \\ 
                                                                                                 & AS      & 229(20)                                                                              & 245(20)                                                                                 & 218(44)                                                                                  & 299(20)                                                                                              & 226(17)                                                                                                   & \textbf{167}(24)                             & 210(16)                                                                                          & 191(15)                                                                                                            & 223                        \\ 
\multirow{2}{*}{PPO}                                                                             & SR      & 27\%(3.4)                                                                            & 52\%(7.8)                                                                                & 41\%(5.8)                                                                                & 26.25\%(6.4)                                                                                       & 52\%(8)                                                                                                   & 44.5\%(4.6)                      & 86.75\%(2.1)                                                                                        & 76\%(5)                                                                                                         & 50.69\%                        \\ 
                                                                                                 & AS      & 262(22)                                                                              & {304}(107)                                                                                 & 521(124)                                                                                 & 241(39)                                                                                           & 372(98)                                                                                                   & 281(87)                          & 240(4)                                                                                          & 293(27)                                                                                                            & 314                        \\ \midrule
\multicolumn{11}{c}{Two-stage method}    \\ \midrule  
\multirow{2}{*}{MOE-VUCA}                                                                        & SR      & 25\%(2)                                                                              & 64\%(8)                                                                                 & 40\%(3.9)                                                                                & 39.5\%(4)                                                                                         & 58\%(9.2)                                                                                                 & 63.25\%(3.3)                     & 86\%(3.6)                                                                                        & 34\%(6.1)                                                                                                          & 51.22\%                        \\ 
                                                                                                 & AS      & \textbf{176}(28)                                                                              & \textbf{194}(23)                                                                                 & \textbf{134}(37)                                                                                  & \textbf{143}(24)                                                                                           & \textbf{161}(46)                                                                                                   & 184(64)                          & \textbf{186}(73)                                                                                          & \textbf{98}(20)                                                                                                             & \textbf{159}                        \\ 

\multirow{2}{*}{\begin{tabular}[c]{@{}c@{}}DM-RCA* \\ \end{tabular}} & SR      & 75\%(1)                                                                              & 81.5\%(3.2)                                                                             & 48\%(2.8)                                                                              & 52\%(7.8)                                                                                         & 85\%(1.7)                                                                                                 & 82.5\%(1.3)                      & 93.5\%(0.8)                                                                                      & 65\%(7.7)                                                                                                          & 72.81\%                        \\ 
                                                                                                 & AS      & 250(15)                                                                              & 214(26)                                                                                 & 276(96)                                                                                  & 263(12)                                                                                           & 237(9)                                                                                                    & 233(66)                          & 204(19)                                                                                          & 230(27)                                                                                                            & 238                        \\ 
\multirow{2}{*}{\begin{tabular}[c]{@{}c@{}}PPO*\\ \end{tabular}}     & SR      & 29.25\%(3.9)                                                                                & 64\%(4)                                                                                    & 35.25\%(5.4)                                                                                    & 42.75\%(6.3)                                                                                           &50.75\%(5)                                                                                                     & 49\%(0.9)                            & 93.75\%(2.1)                                                                                             & \textbf{79.75}\%(4.5)                                                                                                               & 55.56\%                        \\ 
                                                                                                 & AS      & 653(100)                                                                                & 376(80)                                                                                    & 299(50)                                                                                   & 444(67)                                                                                             & 485(88)                                                                                                     & 198(22)                            & 227(4)                                                                                             &226(15)                                                                                                              & 363                        \\ 
\multirow{2}{*}{Ours}                & SR      & \textbf{84.5\%}(1.2)                                                                               & \textbf{83.75}\%(1.7)                                                                                 & \textbf{69.5\%}(1.1)                                                                             & \textbf{56\%}(1.9)                                                                                         & \textbf{88}\%(4.0)                                                                                                   & \textbf{83}\%(3.2)                        & \textbf{95.5}\%(0.8)                                                                                      & 74.5\%(4.3)                                                                                                          & \textbf{79.34}\%                        \\ 
                                                                                                 & AS      & 426(33)                                                                              & 386(37)                                                                                 & 344(22)                                                                                  & 383(56)                                                                                           & 391(50)                                                                                                   & 239(44)                          & 250(24)                                                                                          & 313(18)                                                                                                            & 341                        \\ \bottomrule
\end{tabular}
}

\caption{Model comparison on \emph{Success rate}(SR) and \emph{Arriving step}(AS) using zero-shot scene benchmark, which contains eight modified scenarios. Each scene is a modification of the corresponding scene in few-shot-scene benchmark. The comparison algorithms (DM-RCA and PPO) are One-stage algorithms and MOE-VUCA is a Two-stage training method. For fair comparison, we also use our collected data to pre-train these One-stage methods (denoted by *). The numbers in parentheses are Standard deviation. }
\label{experiment_table}
\end{table*}
\paragraph{Optimization for reasoning.} In order to maintain the stability of the reasoning model, we design a Cumulative-Discount-Reward Weighed loss(DRW-Loss for short) mechanism to optimize the reasoning model asynchronously. The reasoning model is updated at a lower frequency than the PPO. Firstly, we collect the $ \{\mu_{t}^{P},a_{t}^{pred},R_{t}^{nav}\}$ generated by perception and decision-making model during the interaction. Then, we attempt to adopt the $\mu_{t}^{P}$ and $a_{t}^{pred}$ as training samples to optimize the reasoning model. However, at the beginning of training, $\mu_{t}^{P}$ and $a_{t}^{pred}$ are sub-optimal or even incorrect since the policy is not good enough. Such data should be discarded instead of being used to train the reasoning model, as it may cause the model to collapse. Therefore, we calculate the DRW-Loss to stabilize the training. Specifically, DRW-Loss use the \emph{navigation-reward $R^{nav}_{t}$} to calculate the discounted reward $R^{discount}_{t}$. A large discount reward means corresponding action is more valuable. So we use the $R^{discount}_{t}$ to weight the mean squared error corresponding to the sample ${\{\mu_{t}^{P},a_{t}^{pred}\}}$:
\begin{align}
&R_{t}^{discount} = R^{nav}_{t} + \gamma^{1}R^{nav}_{t+1}+...+\gamma^{T-t}R^{nav}_{T}\\
&L^{R}_{1,t} = R_{t}^{discount}*\Vert \mu^{P}_{t}-s^{R}_{t} \Vert_{2}^{2}
\\
&L^{R}_{2,t}=(-1-\log \sigma^{R}_{t}+(\mu^{R}_{t})^{2}+\sigma^{R}_{t})\\
&L^{R}_{t}=L_{1,t}^{R} + \lambda*L^{R}_{2,t}
\end{align}    

where $\gamma,\lambda$ are hyper-parameters. The reasoning model assesses the action generated by the perception and decision-making models through the reward $R_{t}^{sim}$. Meanwhile, the perception and decision-making models produce more pair data $\{\mu_{t}^{P}$, $a_{t}^{P}\}$ with high navigation-reward to promote the evolution of the reasoning model. Two dual learning processes are mutually reinforced by corresponding reward feedback, collectively improving performance.

\section{Experiment}
\subsection{Experiment Setup}

\begin{figure}[h]
	\centering
	\includegraphics[width = 1\linewidth]{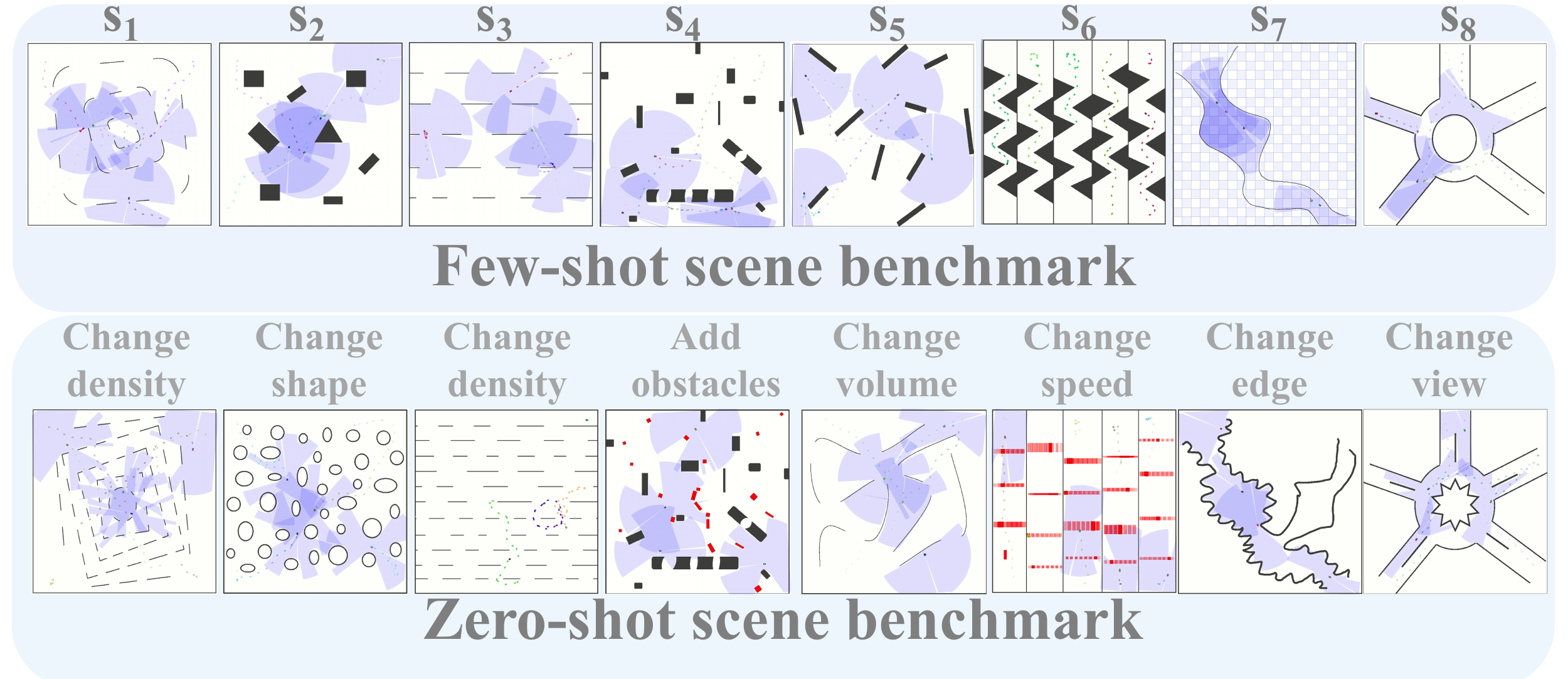}
\caption{Some scenes in two test benchmarks. Each test scenario in the zero-shot scene benchmark is derived from the few-shot scene benchmark, with specific changes.}
	\label{fig:scene_benchmark}
\end{figure}

\paragraph{Scene benchmark.} We adopt four simple scenarios to train the algorithm, including open scene, sparse scene, dense scene, and dynamic scene. To investigate obstacle avoidance and path-finding capabilities, we design two testing benchmarks. As shown in Figure \ref{fig:scene_benchmark}, the first one is few-shot scene, which includes many scenarios similar to training phase. The second is the zero-shot scene, where the scenarios are quite different. We add noise to the observations and adjust the distribution of the goal-position, obstacle-shape, obstacle-speed, and obstacle-density to design zero-shot scene. Zero-shot scene benchmark is used to evaluate the robustness and generalization in new scenarios. We adopt a lightweight robot simulator ROS-Stage\footnote{\url{http://wiki.ros.org/stage}} to implement evaluation, which is more compatible with real robots than OpenAI Gym\footnote{\url{http://gym.openai.com/}}.   


\paragraph{Navigation setting.} The goal point is randomly selected before navigation. The observation is defined as $o_{t}=[o_{t}^{L};o_{t}^{P};o_{t}^{V}]$, which means the concatenation of the Lidar measurement $o_{t}^{L}$, target relative position $o_{t}^{P}$ and robot velocity $o_{t}^{V}$. $o_{t}^{P}$ is relative vectors between the robot's real-time position and the target. The action is defined as $a_{t}=[v_{t};w_{t}]$, representing the Forward and Angular velocity respectively. The observation, latent state and action are all continuous high-dimensional vector. Our reward function has two components, including navigation-reward ($R^{nav}$) and similarity-reward ($R^{sim}$). The $R^{nav}$ is feedback from the simulator, and $R^{sim}$ is calculated by the reasoning model. We design the reward function as follows:
\begin{align}
&R_{t}= R^{sim}_{t}/\rho + R^{nav}_{t}\\
&R^{\text {step}}=\Vert P^{a}_{t}-P^{g}_{t} \Vert_{2}-\Vert P^{a}_{t-1}-P^{g}_{t-1} \Vert_{2} \\
&R_{t}^{\text {nav }}=\left\{
\begin{array}{ll}
R^{\text {goal}}=30, &\text {if reach } \\
R^{\text {collision}}=-20,&\text {if crash } \\
R^{\text {time}}=-0.01,&\text {if alive }\\
R^{\text {step}},&\text {if alive}
\end{array}
\right.
\end{align}

where $P^{a}$ and $P^{g}$ mean the position of the agent and goal, and $\rho$ is a normalized hyperparameter. We set the hyperparameter $\rho$ equal to the probability of a zero vector in a 90-dimensional standard normal distribution.


\paragraph{Dateset collection.}  
Since there is no publicly available navigation dataset, we adopt $A^{*}$ algorithm to plan a global path in a static scene for the first stage of pre-training. Actually, many reasonable strategies can be used to generate the dataset. We have tried alternative methods such as manual annotation, RRT* and finally chose A* as the optimized method for dataset generation. Then we use the Timed Elastic Band (TEB) algorithm to transform trajectory generated by A* into action sequences. The TEB considers the kinematics constraints and optimizes the local action velocity based on the global trajectory. Then, we record observation-action data and manually clean up some exception data. Finally, we collect 200 trajectories, each with about 300 pair data ${\left\{o_{t},a_{t} \right\}}$.

\paragraph{Baselines.} We choose three types of algorithms for comparison. (1) Widely used robot open-source library \textbf{Move-base}\footnote{\url{http://wiki.ros.org/move_base}}. (2) Learning-based navigation algorithms \textbf{DM-RCA}\cite{long2018towards} and \textbf{MOE-VUCA}\cite{vuca}. (3) DRL baseline algorithms \textbf{PPO}\cite{schulman2017proximal} with the same configuration as ours. All comparison algorithms, except the Move-base algorithm are trained in the same scene and each benchmark is evaluated 400 times.

\paragraph{Metrics.}   
The optimization target of the navigation is to avoid collision as much as possible and successfully reach the goal point $o^P_{t}$ in a short time. To achieve this, we adopt two metrics (\emph{Success rate(\%)}, \emph{Arriving step}) to evaluate the performance as previous learning-based works (DM-RCA,MOE-VUCA) did. \emph{Success rate} is a comprehensive indicator of the capability to path-finding and collision avoidance and \emph{Arriving step} assesses the navigation efficiency.

%


\paragraph{Implementation details.} We adopt Microsoft's NNI for tuning parameters in the first stage. In the second stage, we have tested several sets of parameters in a small range and chose the best parameters, and the comparison algorithms also chose the best parameters in the same scene. We set the latent state to 90 dimensions, and the sequence length $n$ is 20, and $\alpha$, $\beta$, $\eta$, $\lambda$, $\gamma$ are 1.0, 20.0, 5e-4, 0.01, 0.99. We discover that updating the reasoning model once every 10 PPO updates is more appropriate. Too frequent updates lead to unstable training, yet too slow updates bring performance degradation as the reasoning model evaluates the action and generates the reward. All models are implemented using PyTorch. We use optimizer Adam with a learning rate of 1e-3 in VAE-Enhanced Demonstration learning and 3e-5 in RL-Enhanced Interaction learning. Video of real-robot experiments are shown at \url{https://youtu.be/jD_7sCdMMWk}. Datasets and part of the code will be released at \url{https://github.com/zwq2018/CL_PDR_NAV}.

\subsection{Results and Analysis}
\begin{figure}[t]
	\centering
	\includegraphics[width = 1\linewidth]{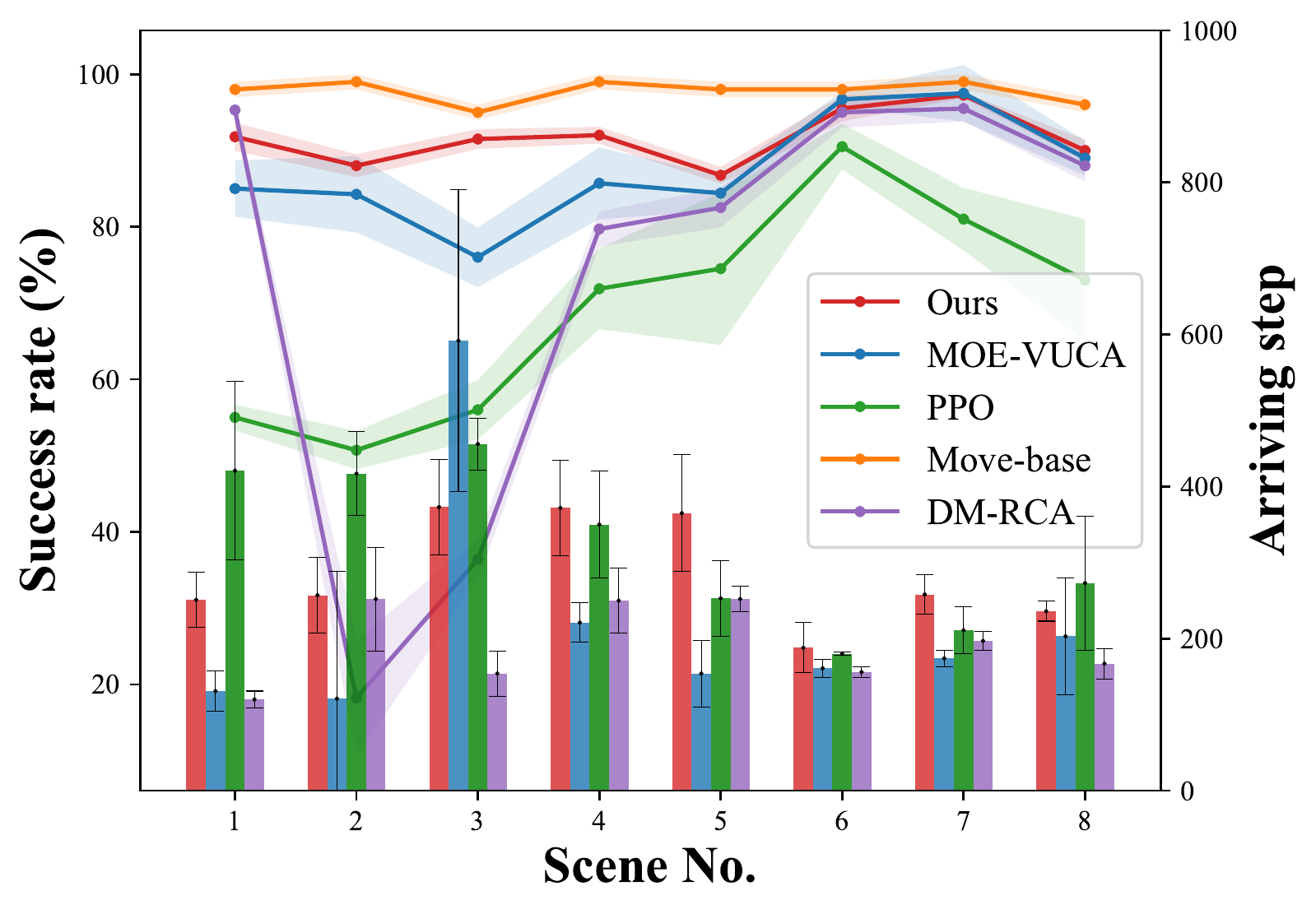}
\caption{Evaluation on few-shot scene benchmark. It includes eight similar scenarios, the line graph represents the \emph{Success rate}(left), and the bar chart means the \emph{Arriving step}(right).}
	\label{fig:experiment_one}
\end{figure}

As Figure \ref{fig:experiment_one}, compared to other methods in few-shot scenes, our algorithm achieves comparable performance as others in most scenarios, except for Move-base. However, Move-base requires an environmental map in advance.  

As shown in Table \ref{experiment_table}, the performance of all the methods in zero-shot scene is significantly degraded. Obviously, environmental distribution shift does have a dramatic impact, and our method is more robust and generalizable than others. In highly dynamic scenes like $S_{6}$-Obstacle change($SC_{6}$ for short), Move-base is almost impossible to avoid the obstacle (10\% in $SC_{4}$ and 15\% in $SC_{6}$), but our algorithm achieves the best results(56\% and 83\%). In complicated maze scenes, our algorithm lead over the runner-up (DM-RCA) by a large margin (9.6\% in $SC_{1}$ and 2.3\% in $SC_{2}$). Additionally, $SC_{8}$ simulates a complex traffic roundabout, and our algorithm is only inferior to PPO-Pretrained method by a small margin(74.5\% compared to 79.75\%). 

MOE-VUCA performs the worst among all methods in zero-shot scenes, although it has the shortest \emph{Arriving step}. We guess it suffers from poor generalization due to the parameter fusion trick it used. DM-RCA is a pretty simple but effective model that greatly outperforms other sophisticated models except ours. It reveals that more sophisticated networks may suffer from the challenge of poor generalization. In addition, we investigate the effect of pre-training using our collected data. For the comparison algorithms, the performance after pre-training with our collected data (denoted by *) does not show a significant improvement over training from scratch. We also observe that the \emph{Arriving step} of ours is much longer than other algorithms. We suspect that the reasoning model encourages the agent to generate more conservative actions for higher similarity-reward $R_{t}^{sim}$, resulting in a longer navigation step. A detailed analysis is provided in Section 4.4. 
\subsection{Ablation Study}
We analyze the contribution of different components in our algorithm. We design three variants of the algorithm: (1)We remove the reasoning model and then train the algorithm using two-stage phases. (2)We replace the DRW-Loss with the mean square error when updating the reasoning model. (3)Without VAE-Enhanced Demonstration Learning phase. Table \ref{ablation_table} illustrates that the reasoning model is essential to our algorithm. The introduction of DRW-Loss and VAE-Enhanced Demonstration Learning can facilitate the learning of the reasoning model.



\begin{figure}[htbp]
\centering

\subfigure[]{
\begin{minipage}[t]{0.49\linewidth}
\flushleft
\includegraphics[width=1\linewidth]{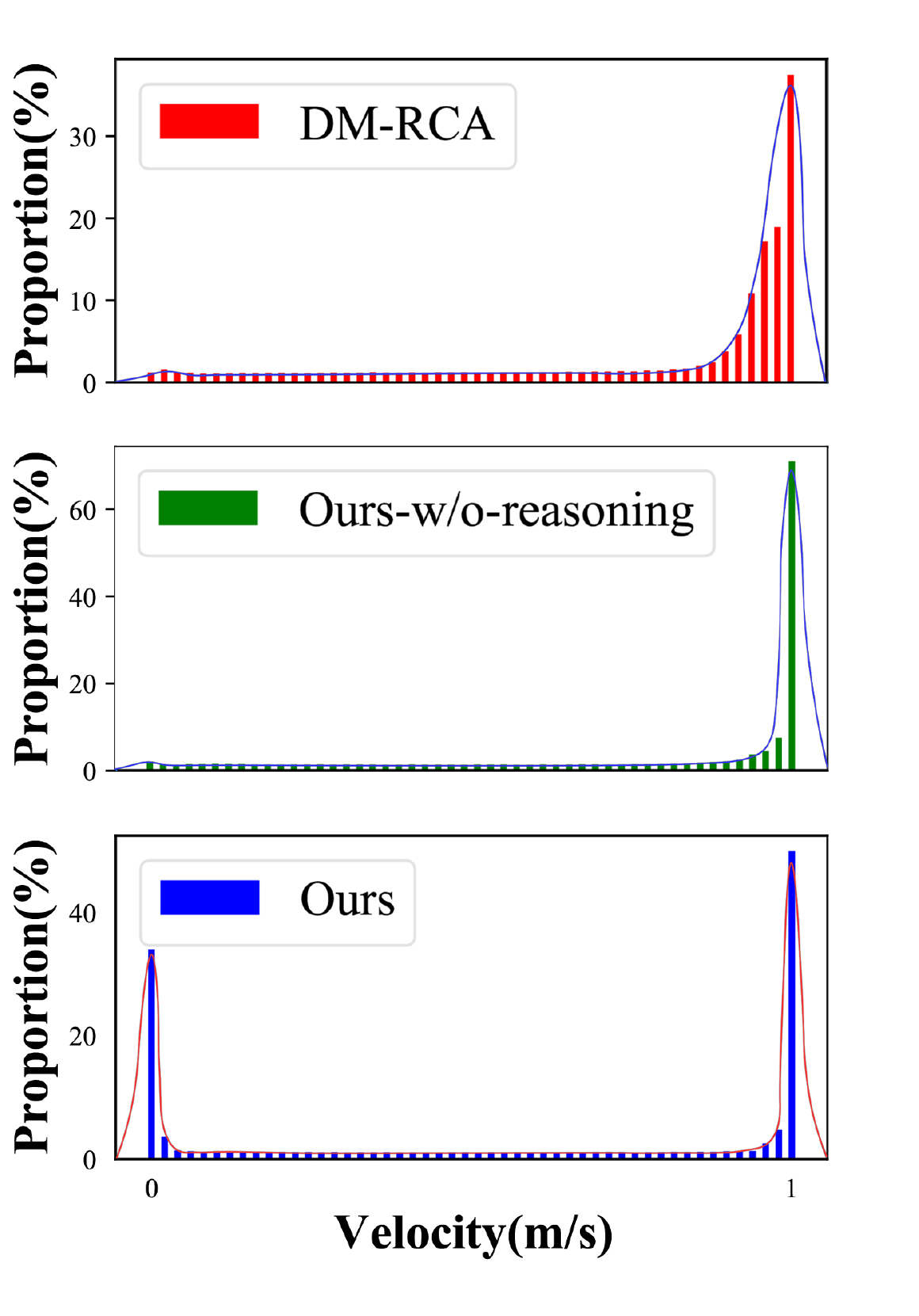}
\label{fig:visual_distribution}
\end{minipage}%
}%
\subfigure[]{
\begin{minipage}[t]{0.49\linewidth}
\centering
\includegraphics[width=1\linewidth]{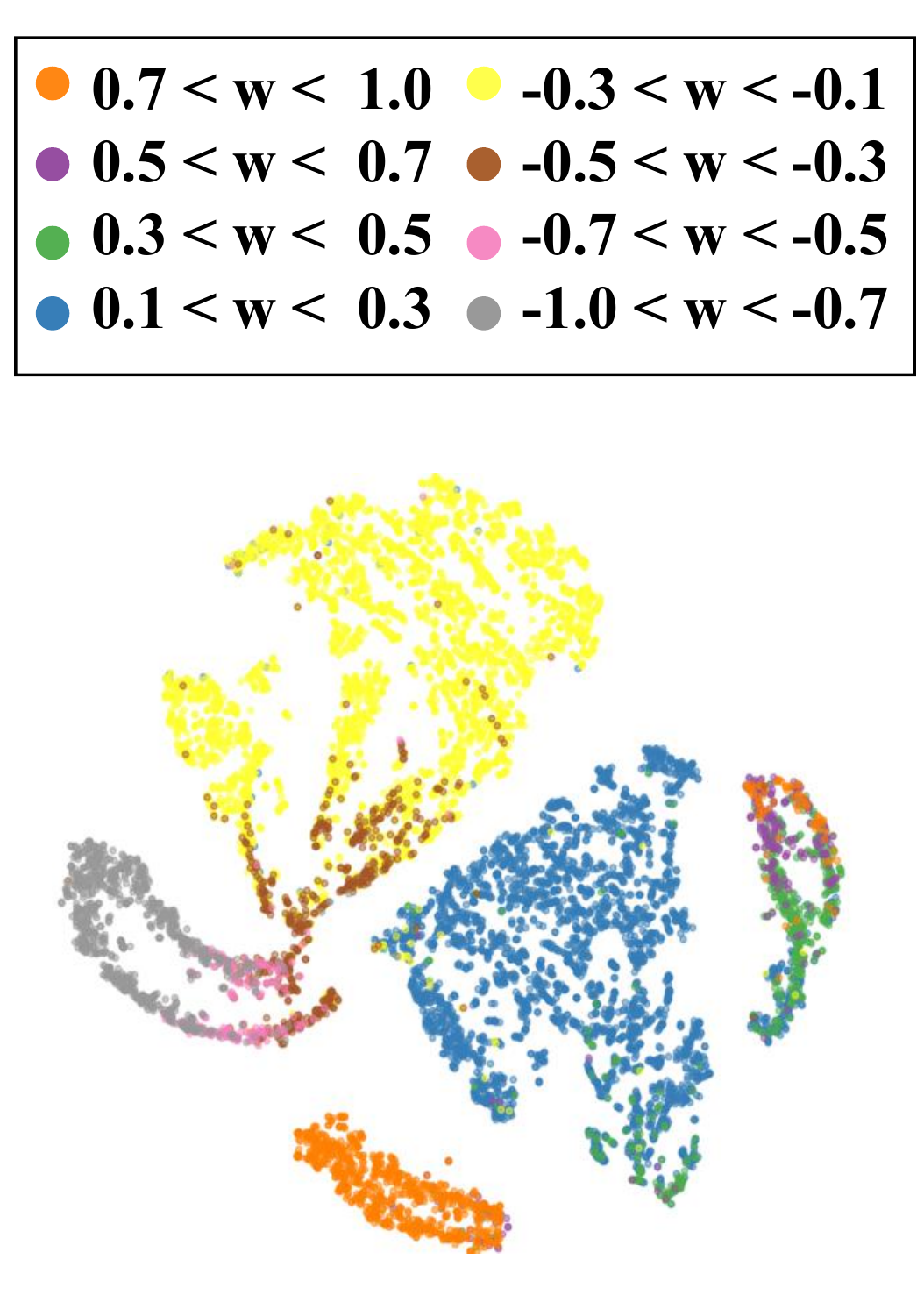}
\label{fig:visual_state}
\end{minipage}%
}%
 
\centering
\caption{(a) The distribution of linear velocity collected by our method shows a bimodal shape. (b) Using t-SEN to visualize the latent states by reasoning model according to different actions. Each point in the figure is a latent state vector after dimension reduction. $w$ means Angular Velocity($rad/s$) of the agent.}

\end{figure}

\begin{table}[H]
\setlength\tabcolsep{2pt}
\begin{tabular}{lr}
\toprule
Model variant                         & Success rate \\
\midrule
Ours                                     &79.3\%        \\
w/o-reasoning model    &55.5\%              \\
w/o-VAE-Enhanced demonstration learning              &63\%      \\
w/o-DRW-Loss                      &55.1\%              \\
\bottomrule
\end{tabular}
\caption{Ablation study on different components(Average).}
\label{ablation_table}
\end{table}

\subsection{Analysis of the Reasoning Model}
We investigate the impact of the reasoning model on actions. The outputs of the decision-making model include Forward Velocity $v$ and Angular Velocity $w$. Firstly, we collect the $w$ and $v$ of several methods in the same scenario and analyze the corresponding behavior patterns. In Figure \ref{fig:visual_distribution}, the action($v$) distribution output by DM-RCA are clustered at a larger velocity value($1m/s$), forming a uni-modal distribution. In contrast, our actions form a bi-modal distribution($0m/s$ and $1m/s$). Thus our strategy is more conservative, preferring to brake rather than bypass when encountering an obstacle. Obviously, comparison policies are more dangerous than ours. It shows that the reasoning model makes the system safer by constraining the action output, albeit sacrificing navigation efficiency. Secondly, we analyze the relationship between the high-dimensional latent vectors $s^{R}$ and the action $a^w$ in Figure \ref{fig:visual_state}. It visualizes the distribution of latent states after dimension reduction. The 90-dim continuous latent states $s^{R}$ are calculated by reasoning model from continuous action sequences $a_{t-n:t}$. These latent vectors are clustered in different regions depending on the action before. It reveals that reasoning model maps similar actions to neighboring regions in latent space, and different actions are projected to distinct. It exhibits a clustering-like effect.


\begin{figure}[t]
	\centering
	\includegraphics[width = 0.9\linewidth]{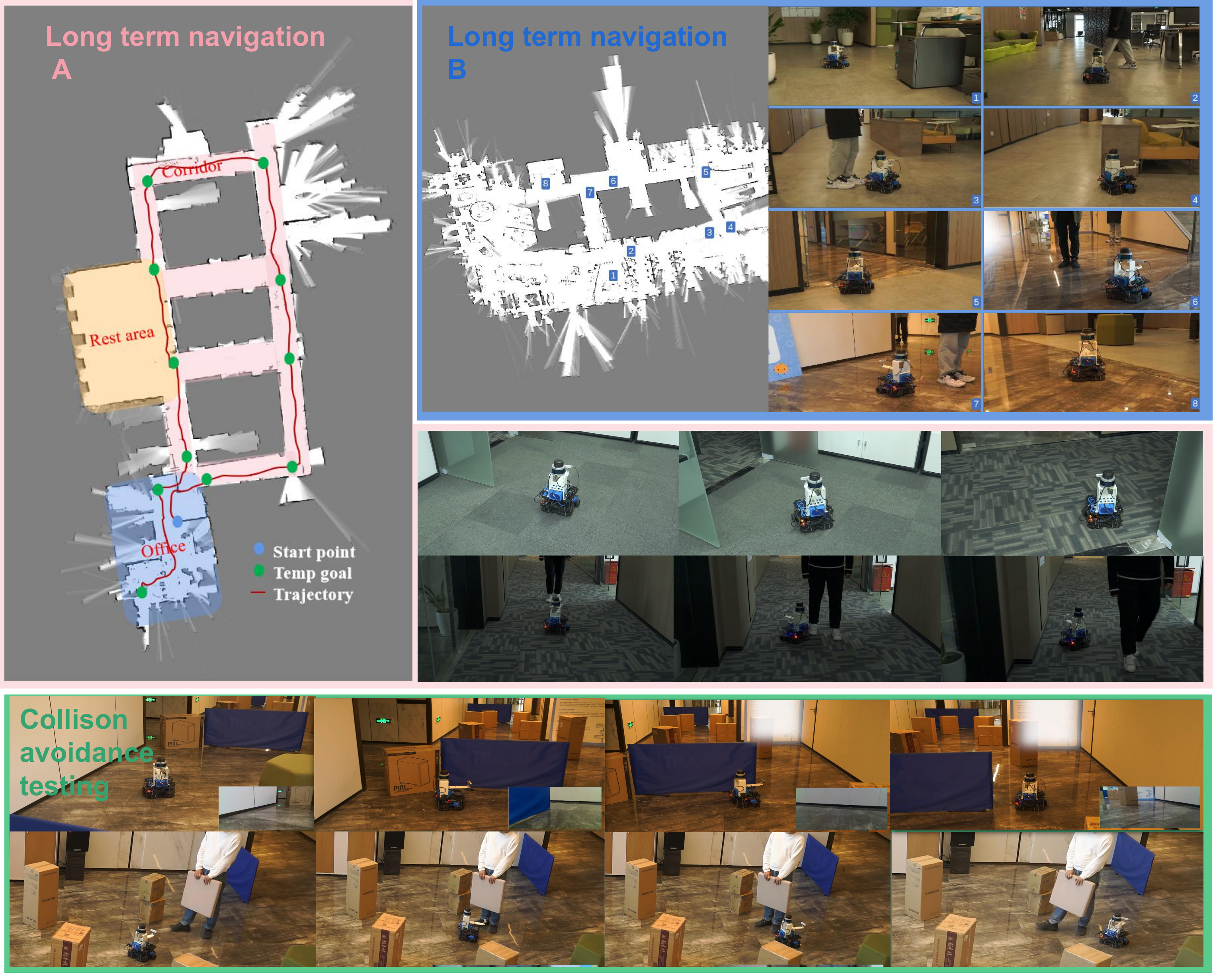}
\caption{Real-world experiments using Turtlebot3 with ROS.}
	\label{fig:real}
\end{figure}

\subsection{Real-World Experiment}
We deploy the algorithm on a wheel robot. As shown in Figure \ref{fig:real}, the robot navigates in a crowded and complex building with moving staff. We randomly select goals and the robot navigates to the targets one by one. The robot is equipped with a 16-line 3D-LIDAR (Velodyne-16), a depth camera (Real-Sense D435) and an edge computing device (NVIDIA Jetson AGX Xavier with Ubuntu and ROS Melodic). All computational processes are performed onboard. Despite the considerable discrepancy between the training scenes and the real scenes, the robot successfully achieve navigation without human help. It means our closed-loop reasoning mechanism can successfully transfer from a simple simulator to a real robot, due to the good adaptability of our approach.



\section{Conclusion}
In this work, we introduce a reasoning process to create an inverse mapping from the output action to the latent state, forming a closed loop with the perception and decision-making process. The reasoning model closes the learning processes of the perception and decision-making and implicitly facilitates the whole system to develop safer and more reasonable collision avoidance behaviors. Experiments have shown that our algorithm is more generalizable and robust in the novel scenario, surpassing the previous approaches in terms of safety and task accomplishment. Real-world robot navigation testing also confirms that its capability being deployed to the real scenes with minimal cost.

\section*{Acknowledgments}
This work is supported by the Key Research and Development Program of Zhejiang Province, China (No. 2021C01013), the National Key Research and Development Project of China (No. 2018AAA0101900), CKCEST, and MOE Engineering Research Center of Digital Library.
\bibliographystyle{named}
\bibliography{ijcai22}
\end{document}